\documentclass[letterpaper]{article} % DO NOT CHANGE THIS
\usepackage{aaai2026}      % remove [submission]
\nocopyright               % hides AAAI copyright box
\usepackage{times}  % DO NOT CHANGE THIS
\usepackage{helvet}  % DO NOT CHANGE THIS
\usepackage{courier}  % DO NOT CHANGE THIS
\usepackage[hyphens]{url}  % DO NOT CHANGE THIS
\usepackage{graphicx} % DO NOT CHANGE THIS
\usepackage{natbib}  % DO NOT CHANGE THIS AND DO NOT ADD ANY OPTIONS TO IT
\usepackage{caption} % DO NOT CHANGE THIS AND DO NOT ADD ANY OPTIONS TO IT
\usepackage{algorithm}
\usepackage{algorithmic}

\usepackage{newfloat}
\usepackage{listings}
\DeclareCaptionStyle{ruled}{labelfont=normalfont,labelsep=colon,strut=off} % DO NOT CHANGE THIS
\floatstyle{ruled}
\newfloat{listing}{tb}{lst}{}
\floatname{listing}{Listing}
\usepackage{xcolor}           % already loaded by most templates

\usepackage{amsmath}
\usepackage{amssymb}
\usepackage{mathtools}
\usepackage{amsthm}

\usepackage{multirow} % For merging cells

\usepackage{subcaption}
\newcommand{\squishlist}{\begin{list}{$\bullet$}{\topsep=1pt \parsep=0pt \itemsep=1pt \leftmargin=1em }} 
\newcommand{\squishend}{\end{list}}

\newcommand{\beitemize}{\begin{list}{$\bullet$}{}} 
\newcommand{\enitemize}{\end{list}}

\theoremstyle{plain}

\theoremstyle{definition}

\theoremstyle{remark}

\title{Mind the Gaps: Auditing and Reducing Group Inequity in Large-Scale Mobility Prediction}
\author{
    Ashwin Kumar\textsuperscript{\rm 1}, 
    Hanyu Zhang\textsuperscript{\rm 2}, 
    David A. Schweidel\textsuperscript{\rm 2}, 
    William Yeoh\textsuperscript{\rm 1}
}
\affiliations{
    \textsuperscript{\rm 1} Washington University in St. Louis, USA\\
    \textsuperscript{\rm 2} Emory University, USA\\
}

\usepackage{bibentry}
\begin{document}

\maketitle

\begin{abstract}
% Next-location prediction is increasingly central to applications in mobility, retail, and public health---but its societal impacts remain underexamined. In this paper, we audit state-of-the-art mobility prediction models trained on a large-scale dataset, and reveal systematic disparities in predictive performance across racial groups. Using aggregate census data, we estimate group-level differences and show that these disparities arise from biases in the underlying data distribution.
Next location prediction underpins a growing number of mobility, retail, and public-health applications, yet its societal impacts remain largely unexplored. In this paper, we audit state-of-the-art mobility prediction models trained on a large-scale dataset, highlighting hidden disparities based on user demographics. Drawing from aggregate census data, we compute the difference in predictive performance on racial and ethnic user groups and show a systematic disparity resulting from the underlying dataset, resulting in large differences in accuracy based on location and user groups.

To address this, we propose \textit{Fairness-Guided Incremental Sampling (FGIS)}, a \textit{group-aware sampling} strategy designed for incremental data collection settings. Because individual-level demographic labels are unavailable, we introduce \textit{Size-Aware K-Means (SAKM)}---a clustering method that partitions users in latent mobility space while enforcing census-derived group proportions. This yields proxy racial labels for the four largest groups in the state: Asian, Black, Hispanic, and White.

Built on these labels, our sampling algorithm prioritizes users based on expected performance gains and current group representation. This method incrementally constructs training datasets that reduce demographic performance gaps while preserving overall accuracy. Our method reduces total disparity between groups by up to 40\% with minimal accuracy trade-offs, as evaluated on a state-of-art MetaPath2Vec model and a transformer-encoder model. Improvements are most significant in early sampling stages, highlighting the potential for fairness-aware strategies to deliver meaningful gains even in low-resource settings.

Our findings expose structural inequities in mobility prediction pipelines and demonstrate how lightweight, data-centric interventions can improve fairness with little added complexity, especially for low-data applications.

\end{abstract}


\section{Introduction}

Next-location prediction has become a central task in applications ranging from mobility planning and retail analytics to public health surveillance. By forecasting where individuals are likely to go, these models support downstream services such as route recommendation, targeted advertising, and resource allocation. However, while accuracy remains the dominant benchmark for evaluating model performance, little attention has been paid to how predictive quality is distributed across different segments of the population.

In this paper, we examine the fairness implications of large-scale mobility prediction. Specifically, we conduct the first comprehensive audit of state-of-the-art next-location prediction models trained on real-world data from millions of users. Our findings reveal consistent disparities in predictive accuracy across racial and ethnic groups, with some groups systematically receiving less accurate predictions than others.

To support fairness analysis and intervention in the absence of individual-level demographic labels, we introduce a novel clustering algorithm called Size-Aware K-Means (SAKM). This method clusters users in latent mobility space while matching target group proportions derived from census data, yielding demographically grounded proxy labels. These clusters enable us to estimate group-level performance metrics and track disparities throughout training.

Building on this foundation, we propose Fairness-Guided Incremental Sampling (FGIS), a lightweight data acquisition strategy for improving equity in low-resource prediction settings. FGIS prioritizes users from underrepresented or underperforming groups during data collection, balancing fairness and accuracy via a tunable tradeoff parameter. Importantly, this intervention operates purely at the data level—requiring no access to user features or modifications to model architecture.

We evaluate our approach using two predictive models: a graph-based MetaPath2Vec model for statewide fairness auditing, and an additional transformer encoder model for controlled intervention experiments in a representative subregion (Tarrant County, Texas, United States). Our results show that FGIS can reduce group disparities by over 40\% in early training stages with minimal impact on final accuracy. These gains are especially pronounced in low-data regimes, highlighting the value of fairness-aware sampling when data is limited or expensive to collect.

\smallskip \noindent \textbf{Contributions:} We summarize our key contributions:
\begin{itemize}
  \item \textbf{Audit at scale:} First large‐scale fairness audit on a SOTA location prediction model with 4.9 million users, uncovering up to 15\% accuracy difference between groups.
  \item \textbf{SAKM proxy labels:} A novel size‐constrained $k$-Means variant that enforces arbitrary census‐derived cluster sizes, enabling demographic fairness evaluation without individual attributes.
  \item \textbf{FGIS sampling:} A plug-and-play batch sampling algorithm that over-samples underperforming groups, reducing early‐stage Total Demographic Parity Violations (TDPV) by over 40\% with minimal overhead.
  \item \textbf{Empirical validation:} Demonstrated across two architectures (MetaPath2Vec and Transformer) and multiple geographies, achieving equity gains with under 1\% long-term accuracy trade-off.
\end{itemize}
% \paragraph{Our contributions are as follows:}
% \begin{itemize}
%     \item We present the first large-scale fairness audit of next-location prediction, revealing persistent disparities in predictive accuracy across geographic, racial and ethnic lines in Texas.
%     \item We introduce \textit{Size-Aware K-Means (SAKM)}, a novel clustering algorithm that respects arbitrary cluster size constraints. We use this to leverage census-derived priors to generate proxy demographic labels for fairness evaluation when individual attributes are unavailable.
%     \item We propose \textit{Fairness-Guided Incremental Sampling (FGIS)}, a practical algorithm for constructing fairer training datasets in streaming or budget-limited settings.
%     \item We empirically demonstrate that FGIS significantly reduces demographic disparities—with over 40\% reduction in group-level performance gaps—while preserving overall model accuracy.
% \end{itemize}

\section{Related Work}

\textbf{Mobility and POI Prediction:}
% \cite{zhang2025colocation}
Recent studies have demonstrated the value of deep learning for point-of-interest (POI) prediction tasks. Transformer-based models have shown strong performance when enriched with auxiliary information, such as travel mode, which helps improve next-location forecasting~\cite{hong2022transformerbaseline}. Other work has focused on the role of routine detection in understanding user behavior, with findings suggesting that consistent travel patterns can inform customer relationship strategies in ridesharing platforms~\cite{dew2024detecting}. To alleviate the challenge of limited real-world data, synthetic datasets such as SynMob have been proposed, offering realistic GPS trajectories for robust model training~\cite{zhu2023synmob}. Additionally, personalized destination prediction has been explored in contextless settings using transformer models trained on partial trajectories~\cite{tsiligkaridis2020personalized}. 
We refer readers to a recent review by \citet{graser2025mobilityreview} on the use of trajectory data for prediction tasks. \citet{zhang2025colocation} present a recent approach to POI prediction, using a colocation network to identify and use similarities between users to more accurately determine visitation patterns. We will use the model from this paper for the bulk of our analyses.

\smallskip \noindent \textbf{Algorithmic Fairness:}
Research on algorithmic fairness has produced a rich taxonomy of group‐ and individual‐level criteria.  Group fairness notions quantify disparities in aggregate error or allocation rates across protected groups, with \emph{Demographic Parity} (also called Statistical Parity) being one of the earliest and most widely adopted definitions.  A predictor satisfies demographic parity when its positive‐prediction rate is identical for every group~\cite{DP_dwork2012}.  Follow-up work proposed alternative group metrics—most notably \emph{equalized odds} and \emph{equality of opportunity}, which require parity of error rates or true-positive rates conditional on the ground truth~\cite{hardt2016equality}.  Complementary research advocates \emph{individual fairness}, urging that ``similar individuals be treated similarly,'' though this is difficult to enforce when similarity measures are ill-defined.

Fairness goals are shaped by modeling assumptions and data limitations~\cite{mitchell2021algorithmic,barocas2023fairness}. Demographic parity remains appealing in settings like mobility prediction, where ground-truth labels are limited, as it depends only on observed prediction disparities. We use demographic parity as a diagnostic tool, adapting it to the multi-group setting.
% \ak{Add refs for TDPV like metrics}

\smallskip \noindent \textbf{Fairness in Active Learning:}
Fair active learning aims to select data points for labeling in a way that enhances model fairness. Early work in this area focused on balancing label acquisition to satisfy group fairness constraints in supervised settings~\cite{shekhar2021adaptive}. More recent studies have extended this to streaming scenarios~\cite{wang2023preventing} and developed sampling methods that promote fairness without requiring group labels during training~\cite{pang2024fairness}.
Our setting differs significantly, as we treat it as a data acquisition problem, rather than a labeling problem. We do not assume access to user features (e.g., location traces) at selection time. Instead, we assume a pool of users, each associated with a known group label, and develop a strategy for sampling users in a way that promotes fairness in data acquisition. Our approach is feature-agnostic and focuses on balancing data representation across groups to improve model equity.

\smallskip \noindent \textbf{Fairness in Mobility Prediction:}
Fairness concerns have only recently reached spatio-temporal modeling.  Early efforts focused on equitable demand prediction for ride-hailing and public-transit systems~\cite{yan2019fairst, kumar2023SI, zheng2023fairness}.
Beyond demand forecasting and ridehailing, POI recommendation studies have proposed fairness metrics to ensure balanced exposure of venues or user segments. For instance,~~\citet{weydemann2019defining} introduced utility- and diversity-based fairness criteria to mitigate popularity bias and ensure equitable treatment across user demographics. More recent work has begun to examine whether predictive systems systematically underperform for marginalized communities~~\cite{zheng2021equality, zhang2024travel}, highlighting the need for algorithmic interventions that can reduce error disparities without compromising accuracy.

%\vspace{1mm}
\smallskip \noindent \textbf{Summary:}
Our paper differs from prior mobility fairness research in two key ways. First, we audit \emph{individual-level next-location prediction}, in contrast to previous work that focuses on region-level demand or recommendation outcomes. This allows us to reveal disparities that persist even over short horizons and individual trajectories. Second, instead of applying fairness regularizers or post-hoc adjustments, we propose a \emph{data acquisition} strategy that improves demographic parity with minimal impact on predictive performance—complementing model-side fairness interventions in the literature.

\section{Background: Problem Setting}

We now provide some background on the prediction task of interest as well as the disparity measurement used.

\subsection{Prediction Task}

Our paper focuses on evaluating the fairness of next-location prediction models trained on large-scale mobility data. The underlying task is to forecast where a user will go next based on their historical movement patterns. Formally, each user's trajectory is represented as a time-ordered sequence of visits to points of interest (POIs), where each visit is encoded as a POI identifier and a timestamp.

Given such a sequence, the model is trained to predict the user's next POI. This task is inherently challenging due to the wide variability in user behavior, the heterogeneity of POIs across regions, and the complex temporal dynamics of human mobility. Moreover, mobility patterns are shaped by socioeconomic and spatial factors, which may result in uneven model performance across different user groups.

To evaluate predictive performance, we use the \textbf{top-$k$ accuracy metric} over a fixed future time window. In our analysis, we adopt a \textbf{one-week lookahead period}, and report the \textbf{1-week Acc@20} metric. This measures the fraction of test instances (users) for which at least one of the user’s actual future POI visits during the next week appears in the top-20 predictions generated by the model. 

This formulation allows us to quantify how well the model anticipates user behavior at a practically meaningful granularity. Importantly, it also enables disaggregated evaluation by demographic group, allowing us to assess whether certain populations systematically experience lower prediction accuracy, which forms the basis of our fairness audit.

\subsection{Disparity Measurement}

% \memo{This should go to the Fairness Analysis section?}

We consider group fairness in our evaluations, where we use a modified version of demographic parity~\cite{DP_dwork2012, hardt2016equality} called \emph{Total Demographic Parity Violations (TDPV)}: 
\begin{align}
    \text{TDPV} = \sum_{i < j} \left| z_{g_i} - z_{g_j} \right|
\end{align}
Here, \( z_{g} \) denotes the prediction accuracy for group \( g \), measured as the average top-$k$ prediction accuracy in that group. This metric captures the total disparity in performance across demographic groups by summing the absolute pairwise differences in accuracy. A lower TDPV indicates more equitable accuracy distribution, while a higher TDPV signals that some groups experience significantly better or worse performance than others. The metric is symmetric and unweighted, treating all group pairs equally regardless of their population sizes.

\section{Fairness Analysis}

We aim to determine whether next-location prediction models trained on large-scale mobility data exhibit systematic disparities in performance across racial and ethnic groups. Our audit examines whether unequal prediction accuracy may arise from spatial bias, imbalanced data coverage, or overfitting to overrepresented populations. We describe the various components of our analysis below.

\subsection{Model: MetaPath2Vec}

To evaluate fairness in a realistic predictive setting, we use the MetaPath2Vec model~\cite{zhang2025colocation}, a state-of-the-art approach for next-location prediction in heterogeneous networks. The model operates over a bipartite user-POI graph, where nodes represent users and POIs, and edges represent observed visits.

MetaPath2Vec performs random walks guided by predefined meta-path schemas (e.g., user-POI-user) to generate sequences of nodes, which are used in a skip-gram objective~\cite{dong2017metapath} to learn latent embeddings. This enables the model to position each user close to visited and structurally similar POIs in the embedding space. These embeddings are then used to predict the next POI a user is likely to visit.

% % Hanyu's original text
% The goal of network representation learning is to embed nodes such that proximity in the latent space reflects structural similarity in the original network. Considering heterogeneous
% networks with different entities and relations, meta-path based random walks (i.e., MetaPath2Vec) offer an efficient means of transforming the heterogeneous network structure into the skip-gram model~\cite{dong2017metapath}. In a bipartite network with users and POIs as nodes in
% two disjoint sets, it is assumed users can visit POIs shown by the links between users and POIs. Therefore, MetaPath2Vec constructs node sequences by traversing the network according to predefined meta-path schemas, such as user-POI, which encode typical interaction pathways ~\cite{zhang2025colocation}. These sequences allow the model to learn embeddings that position each user close to POIs they have visited and to structurally similar POIs. By capturing these relational patterns, MetaPath2Vec supports accurate prediction of a user’s next likely location based on their embedding in the learned space.

\subsection{Texas Mobility Dataset}

We evaluate the model on mobility data collected from mobile devices detected in Texas between January 1 and April 15, 2021. This dataset was sourced from a third-party location analytics firm and collected passively via mobile applications, depending on user permissions. The raw dataset contains data point including a hashed device ID, timestamp, and GPS coordinates. Pings matched to known POIs also include metadata such as name, category, and address.

From this raw stream, we constructed anonymized trajectories for approximately 4.9 million users, comprising ordered sequences of visits to over 530{,}000 unique POIs. Each user trajectory varies in length from 1 to 2{,}625 visits, with an average of 67. Home locations are inferred using nighttime GPS activity and assigned to ZIP Code Tabulation Areas (ZCTAs); no exact coordinates or personal identifiers are retained.

We use results from a pre-trained MetaPath2Vec model trained on user data from this anonymized Texas dataset in our analysis.\footnote{The processed trajectory dataset and the trained model's results were obtained with permission from the original authors~\cite{zhang2025colocation}. We are unable to share the dataset publicly due to privacy constraints.}

% Original: Hanyu's text
% Our data were provided by a third-party location analytics firm that passively collects GPS signals through a network of mobile apps, depending on app permissions and installation. We were given access to detailed location data for all mobile devices detected in Texas between January 1 and April 15, 2021. The raw data contained a hashed device identifier, timestamps, and GPS coordinates (latitude and longitude). For pings occurring at tagged points of interest (POIs), the provider appended metadata such as the name, category, and address of the establishment. 

% We constructed anonymized mobility trajectories for approximately 4.9 million users, comprising time-ordered sequences of visits to over 530,000 unique POIs. Each visit is represented by a POI identifier and a corresponding timestamp. Trajectory lengths vary widely across users, ranging from 1 to 2{,}625 visits, with an average length of approximately 67 visits per user.

% Approximate home locations are inferred from aggregated nighttime GPS activity and used to assign users to ZIP Code Tabulation Areas (ZCTAs); no precise coordinates or personally identifying information are retained.

\subsection{Demographic Data and Motivation}

Since individual demographic attributes are not available in the dataset, we infer coarse group membership using publicly available census statistics. This is essential for evaluating whether the model yields unequal outcomes across demographic groups despite being trained without such labels.

We focus on the four largest racial and ethnic groups in Texas—Hispanic or Latino, White (non-Hispanic), Black or African American (non-Hispanic), and Asian (non-Hispanic)—which together account for 96.1\% of the population~\cite{uscb2023props} (see Table~\ref{tab:texas_demographics_percent}).

Using census-reported racial composition at the ZCTA level, we estimate a probability distribution over each user's group membership based on their home ZCTA.
% Each user is probabilistically assigned to these groups based on the demographic distribution of their home ZCTA, as inferred from GPS data. 
We also perform complementary analysis at the county level, using the same inference procedure. We compute fairness metrics at both ZCTA and county resolution to examine whether disparities persist across geographic scales. These inferred probabilities are used solely in aggregate form to evaluate fairness, and are never treated as ground-truth labels.

\begin{table}[t]
    \centering \small
    \caption{Texas Population by Race and Ethnicity}
    \begin{tabular}{|l|c|}
        \hline
        \textbf{Demographic Group} & \textbf{Percentage (\%)} \\
        \hline
        Hispanic or Latino (any race)        & 39.8 \\
        White alone (not Hispanic)           & 38.7 \\
        Black or African American alone      & 12.0 \\
        Asian alone                          & 5.6 \\
        \hline
    \end{tabular}
    \label{tab:texas_demographics_percent}
\end{table}

\subsection{Evaluation Method}

To estimate group-level disparities in predictive performance, we compute top-20 prediction accuracy within a one-week lookahead window (1-week Acc@20). A prediction is considered correct if any POI visited in the next week appears in the model’s top-20 ranked list.

Because we lack individual-level race/ethnicity labels, we adopt two common assumptions:
\begin{enumerate}
    \item \textbf{Geographic representativeness:} Users in each region (ZCTA or county) are treated as a random sample of that region’s population.
    \item \textbf{Intra-region uniformity:} Prediction accuracy is assumed constant across demographic groups within a region.
\end{enumerate}

\subsubsection{Group-Level Accuracy ($z_g$) Computation:}
To estimate prediction accuracy for each demographic group, we aggregate model performance over geographic regions using census-based priors. For each region $r$, we compute the average top-20 prediction accuracy $a_r$ based on all users assigned to that region.
% We assume that prediction performance is approximately uniform across individuals within a region.

Let $n_r$ denote the number of users in region $r$ in the dataset, and let $p_{g,r}$ be the proportion of group $g$ in that region according to census data. We estimate the number of correct predictions attributable to group $g$ in region $r$ as
%\begin{align}
    $c_{g,r} = a_r \cdot n_r \cdot p_{g,r}$.
%\end{align}
The total number of correct predictions and total population for group $g$ are then
%\begin{align}
    $C_g = \sum_r c_{g,r} = \sum_r a_r \cdot n_r \cdot p_{g,r}$ and 
    $N_g = \sum_r n_r \cdot p_{g,r}$, respectively.
%\end{align}
The group-level accuracy is then defined as
%\begin{align}
    $z_g = \frac{C_g}{N_g}$, 
%\end{align}
which represents the expected prediction accuracy experienced by a typical member of group $g$, assuming geographically uniform accuracy within each region. The resulting $z_g$ values are used to compute fairness metrics in our analysis.

\section{Fairness Audit: Observed Disparities}

Using the trained MetaPath2Vec model, we evaluate next-location predictions for all users based on the 1-week Acc@20 metric. 
To examine variation in model performance, we assess disparities along the race/ethnicity axis.
% To examine variation in model performance, we assess disparities along two dimensions: geography and race/ethnicity.

\subsection{Racial and Ethnic Disparities}
% We next assess disparities in predictive performance across racial and ethnic groups. 
Using the group-level estimation procedure described earlier, we compute the expected accuracy experienced by a typical member of each demographic group. Table~\ref{fig:mean_racial} reports the mean Acc@20 by group at both ZCTA and county levels.
% In addition, Figure~\ref{fig:kde} presents kernel density estimates of the accuracy distributions across the population of each group, showing the density of users at each accuracy level.

Our analysis reveals that White users are expected to experience the highest accuracy, followed by Hispanic, Asian, and Black users. This disparity is more pronounced at the ZCTA level, while county-level aggregation tends to smooth out local differences. Nevertheless, the observed gap in performance across groups persists at both geographic resolutions, suggesting systemic disparities in the model’s predictive behavior. These disparities may reflect uneven data distribution, differential mobility patterns, or structural biases learned during training. Additional results measuring geographical variations are included in the supplement.

\begin{table}[t]
    \centering \small
    \caption{Mean Acc@20 by Group and Region, including Total Demographic Parity Violations (TDPV)}
    \begin{tabular}{|l|c|c|}
    \hline
    \textbf{Group} & \textbf{ZCTA} & \textbf{County} \\
    \hline
    White   & \textbf{0.390} & \textbf{0.383} \\
    Hispanic & 0.355 & 0.359 \\
    Asian   & 0.346 & 0.353 \\
    Black   & 0.335 & 0.351 \\
    \hline
    \hline
    \textbf{TDPV} & 0.174 & 0.102 \\
    \hline
    \end{tabular}
    \label{fig:mean_racial}
\end{table}

\section{Group-Aware Incremental Sampling}

Having identified a clear bias in the model predictions, we look at a solution to mitigate disparities in next-location prediction. We consider a practical setting in which model developers acquire mobility data over time, subject to budget constraints. Suppose there exists a large population of potential users, each with a hidden trajectory history, from which a learning agent incrementally samples training data. Our goal is to actively guide this sampling process to improve fairness across demographic groups, by constructing training sets that yield more equitable predictive performance.

This strategy requires some notion of group membership for each user. While individual-level demographic attributes are not observed, we assume access to coarse-grained proxy labels based on the user’s home region. These proxies are inferred using publicly available census data and are used solely in aggregate form to steer the sampling process.

Let $\mathcal{U}$ denote the full user population and let $\mathcal{D}_t \subset \mathcal{U}$ be the training set at iteration $t$, initialized as $\mathcal{D}_0 = \emptyset$. At each round $t = 1, \dots, n$, the agent selects a batch of $B$ users from $\mathcal{U} \setminus \mathcal{D}_{t-1}$, optionally conditioned on their (inferred) group, and adds them to the dataset: $\mathcal{D}_t = \mathcal{D}_{t-1} \cup \mathcal{S}_t$, where $|\mathcal{S}_t| = B$. A predictive model $\mathcal{M}_t$ is trained on $\mathcal{D}_t$ and evaluated using top-$k$ accuracy. For each group $g$, we compute the group-specific accuracy $z_g^{(t)} = \mathrm{Acc}_k(\mathcal{M}_t \mid g)$. Note that in this model, the full set of user features is only acquired after we decide to sample them.

We aim to guide the sequence of samples $\{\mathcal{S}_t\}_{t=1}^n$ to reduce disparity across the $z_g^{(t)}$ metrics, by adaptively prioritizing users from groups that are underrepresented or underperforming. 
% Crucially, this requires a meaningful assignment of users to demographic groups. Rather than assigning users arbitrarily or uniformly, we seek a proxy labeling that reflects the census-reported racial distribution of the region. This motivates the use of a clustering method that can produce demographically grounded group assignments aligned with population priors.
% To this end, we introduce a modified clustering algorithm—Size-Aware K-Means (SAKM)—which partitions users into groups while matching the target racial proportions observed at the ZCTA level. SAKM enables principled construction of proxy group labels that support fairness-aware learning, described next.
To support this group-aware sampling, we require user-level group assignments that reflect the demographic makeup of the population. Since individual attributes are not observed, we turn to unsupervised clustering to assign users to demographic groups in a principled way. Our goal is to generate proxy labels that align with census-reported racial proportions at the regional level. This leads us to a constrained clustering approach that incorporates group size targets into the clustering objective, which we describe next.%We describe this method, \emph{Size-Aware K-Means (SAKM)}, in the supplement.

\subsection{Size-Aware K-Means (SAKM)}

We implement a modified clustering algorithm, \textit{Size-Aware K-Means (SAKM)}, which extends standard $k$-means to enforce user-defined cluster size constraints. Given a target group distribution $\boldsymbol{\pi} = (\pi_1, \dots, \pi_k)$, SAKM aims to produce clusters of sizes close to $\pi_g \cdot N$, where $N$ is the total number of users.

SAKM introduces a Lagrangian penalty in the assignment step, so the cost of assigning a point $x_i$ to cluster $g$ becomes:
\begin{align}
    \text{cost}(x_i, g) = \|x_i - \mu_g\|^2 + \lambda_g \label{eq:sakm-cost}
\end{align}
where $\mu_g$ is the centroid of cluster $g$ and $\lambda_g$ is a Lagrange multiplier that penalizes deviations from the target size. These multipliers are updated iteratively as:
\begin{align}
\lambda_g \leftarrow \lambda_g + \eta \cdot \left( \frac{n_g}{N} - \pi_g \right) \label{eq:sakm-largrange}
\end{align}
where $n_g$ is the current number of points in cluster $g$. 
To resolve the label ambiguity and improve convergence quality, we run the optimization over all permutations of the target proportions $\boldsymbol{\pi}$ and select the clustering with the lowest objective (inertia). This permutation search is motivated by a key challenge in centroid initialization: if a randomly initialized centroid is far from the true region of the corresponding size, the resulting cluster may fail to attract the intended mass of points. By evaluating all $k!$ permutations of the size targets, SAKM increases the likelihood that size constraints are matched with semantically meaningful partitions. While the worst-case complexity grows factorially with the number of groups, this remains tractable in our setting where $k=4$. We provide more details on SAKM, including its pseudocode and calibration results, in the supplement.

%For larger $k$, the cost could be reduced using combinatorial assignment methods such as the Hungarian algorithm with $\mathcal{O}(k^3)$ complexity, which we explored but do not report in this paper. 

\subsection{Fairness-Guided Sampling Strategy (FGIS)}

Given the proxy group assignments produced by SAKM, we now seek to actively construct training datasets that reduce disparities in model performance across groups. At each iteration, we select new users from the population based on the expected impact their group membership will have on fairness outcomes.

Our strategy for \textit{Fairness Guided Incremental Sampling (FGIS)} is based on the following intuition: additional data improves group-level accuracy $z_g$, but with diminishing returns---each new user contributes less than the last. Moreover, not all groups benefit the same from additional samples: improving underrepresented or underperforming groups will offer greater marginal gains in fairness. We therefore design a sampling rule that assigns higher weight to groups expected to most improve performance parity, by increasing the weights of groups with lower data representation and lower accuracy.  

Recall that $z_g$ denotes the top-$k$ accuracy for group $g$ under the current model, and let $x_g$ denote the number of users from group $g$ currently included in the training set. We define the sampling weight for group $g$ as:
\begin{align}
w_g \propto \left[ z_g \cdot (x_g + 1) \right]^{-\beta} \label{eq:sampling-weight}
\end{align}
where $\beta \in [0,\infty)$ is a tunable parameter controlling the trade-off between uniform sampling ($\beta = 0$) and fairness-aware sampling ($\beta > 0$). This form reflects a first-order approximation of the expected fairness gain from sampling group $g$: groups with low accuracy $z_g$ and few seen users $x_g$ are prioritized, while groups that already perform well or have large training representation are de-emphasized.

\begin{algorithm}[t]
\small
\caption{Fairness-Guided Incremental Sampling Loop}
\label{alg:sampling_main}
\textbf{Input}: Full user set $\mathcal{U}$, proxy group labels $g(u) \in \{1, \dots, G\}$, batch size $B$, rounds $n$\\
\textbf{Parameter}: Sampling trade-off $\beta$, initial accuracy estimate $z_g^{(0)} = 0.1$\\
\textbf{Output}: Accuracy metrics $\{z_g^{(t)}\}$ for $t = 1 \dots n$
\begin{algorithmic}[1]
\STATE Initialize dataset $\mathcal{D}_0 \gets \emptyset$; seen user counts $x_g \gets 0$ for all $g$
\STATE Set accuracy estimates $z_g^{(0)} \gets 0.1$ for all $g$
\FOR{$t = 1$ to $n$}
    \STATE $\mathcal{S}_t \gets$ \textsc{Sample}($\mathcal{U} \setminus \mathcal{D}_{t-1}, \{x_g\}, \{z_g^{(t-1)}\}, B, \beta$)
    \STATE $\mathcal{D}_t \gets \mathcal{D}_{t-1} \cup \mathcal{S}_t$
    \STATE Train model $\mathcal{M}_t$ on $\mathcal{D}_t$
    \STATE Evaluate $\mathcal{M}_t$ to obtain group accuracies $\{z_g^{(t)}\}$
    \STATE Update $x_g \gets x_g + \#\{u \in \mathcal{S}_t : g(u) = g\}$
\ENDFOR
\STATE \textbf{return} accuracy metrics $\{z_g^{(t)}\}$
\end{algorithmic}
\end{algorithm}

\begin{algorithm}[t]
\small
\caption{\textsc{Sample}: Fairness-Aware Batch Selection}
\label{alg:sample}
\textbf{Input}: Candidate users $\mathcal{C}$ with group labels $g(u)$,\\
\hspace{1.9em} Group counts $\{x_g\}$, group accuracies $\{z_g\}$, batch size $B$, mini-batch size $m$, trade-off $\beta$\\
\textbf{Output}: Sampled batch $\mathcal{S}_t$ of $B$ users
\begin{algorithmic}[1]
\STATE Initialize sampled set $\mathcal{S}_t \gets \emptyset$
\WHILE{$|\mathcal{S}_t| < B$}
    \STATE Compute group weights $w_g \propto \left[ z_g \cdot (x_g + 1) \right]^{-\beta}$
    \STATE Normalize $\{w_g\}$ to form group-level distribution $p_g$
    \STATE Assign each user $u \in \mathcal{C} \setminus \mathcal{S}_t$ probability $p_{g(u)}$
    \STATE Sample $m$ users from $\mathcal{C} \setminus \mathcal{S}_t$ using these probabilities
    \STATE Add sampled users to $\mathcal{S}_t$; update counts $x_g$ accordingly
\ENDWHILE
\STATE \textbf{return} sampled users $\mathcal{S}_t$
\end{algorithmic}
\end{algorithm}

% Ideally, we would sample a single user and retrain the model, but this becomes computationally prohibitive. We instead sample a batch of $B$ users. The weight $w_g$ depends on $x_g$, which will change for each additional user we sample. So, even within a single batch, we need to sample one user at a time, update $x_g$ and recompute weights $w_g$ repeatedly to sample accurately. In practice, we observe that sampling minibatches of users leads to similar results, as long as the minibatch size $m$ is relatively small. We use $m=50$, $B=1000$ for our experiments.
Ideally, we would sample a single user, retrain the model, and update the group accuracy estimates before sampling the next. This would allow the weights $w_g$ to reflect the most up-to-date performance information. However, retraining after every user is computationally prohibitive. As a first simplification, we instead train the model once per batch of $B$ users.

Even within a single batch, the weights $w_g$ depend on $x_g$, which changes as users are added to the training set. To sample accurately under this dependency, we would need to recompute $w_g$ after each individual selection. As a second simplification, we instead sample users in mini-batches of size $m$, updating $x_g$ and $w_g$ after each mini-batch rather than after each user. In practice, we find that this approximation performs comparably when $m$ is small. We use $m = 50$ and $B = 1000$ in our experiments.
 
% \ak{[Optional: insert alternate derivation of this formula here, depending on whether we keep the Nash Welfare justification or replace it with a new fairness objective.]}

To implement this, we maintain per-group accuracy estimates and user counts over sampling iterations. At each sampling step, we compute group weights using the formula above, map weights to per-user probabilities, and select a small mini-batch of users to add to the training set. The procedure repeats until the batch budget is exhausted. We use log-domain computation to maintain numerical stability and re-normalize probabilities after each mini-batch. Algorithms~\ref{alg:sampling_main} and~\ref{alg:sample} outline this process.

This iterative, mini-batch design ensures that sampling remains responsive to updated accuracy estimates as the model improves. The resulting datasets reflect a data-efficient path toward performance parity, trading off global representativeness for reduced inter-group variance in a controlled manner.

% \paragraph{Effect of $\beta$:} Intuitively, $\beta$
% \ak{ subsection{Evolution with beta}
 % As $\beta\rightarrow$ 1, is inversely prop to z and count. For larger beta, it quickly becomes: sample mostly from smaller group}

\section{Experimental Results}

We now describe the results of our empirical evaluations.

\subsection{Experimental Setup}

To evaluate our proposed approach, we train models on the location dataset with a variety of $\beta$ values, showing how this parameter changes the utility-fairness tradeoff. Since retraining on the entire Texas dataset for multiple $\beta$ and getting confidence intervals is prohibitive, we select \textbf{Tarrant County} as a representative region within Texas. Further, to evaluate the effectiveness of our approach beyond the MetaPath2Vec model, and to also measure the relative strength of MetaPath2Vec, we additionally train a transformer encoder based model for location prediction, based on work on using transformers for mobility prediction~\cite{hong2022transformerbaseline}. The selection criteria for Tarrant County and model details, hyperparameters, and other details about the experimental setup are included in the supplement.

\begin{figure}[t]
  \centering
  % (a) Transformer
  \begin{subfigure}{\columnwidth}
    \centering
    \includegraphics[width=0.49\linewidth]{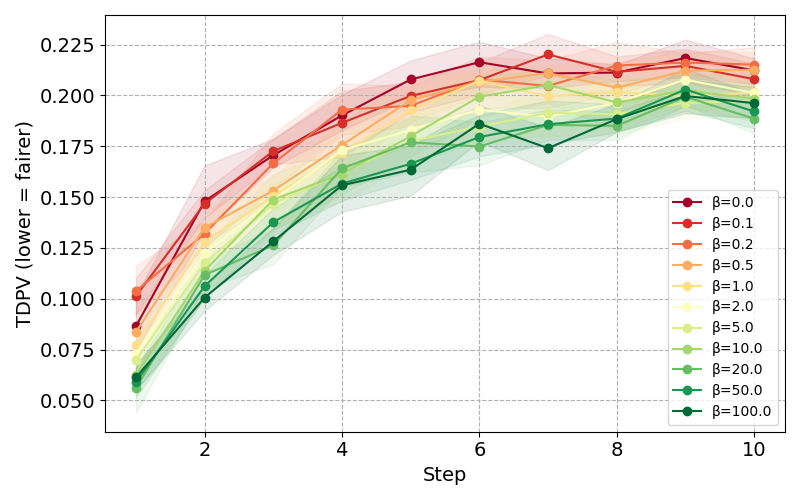}%
    \includegraphics[width=0.49\linewidth]{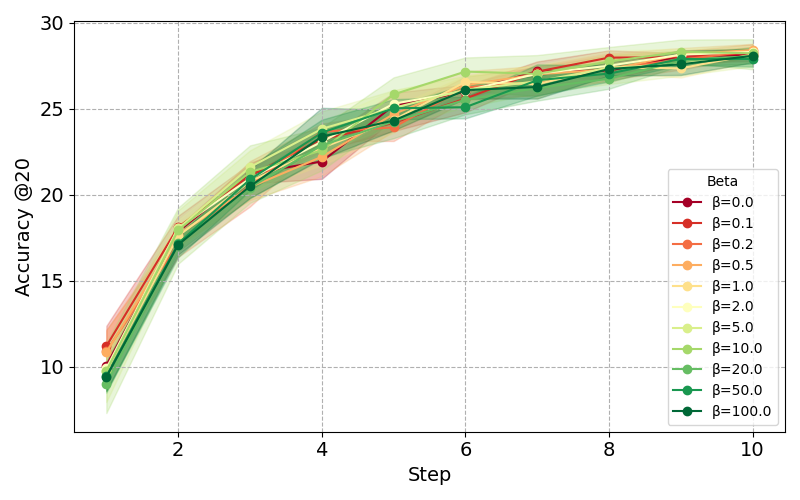}
    \vspace{-0.5em}
    \caption{Transformer}
    \label{fig:results_vs_step_transformer}
  \end{subfigure}

  \vspace{1em}

  % (b) MetaPath2Vec
  \begin{subfigure}{\columnwidth}
    \centering
    \includegraphics[width=0.49\linewidth]{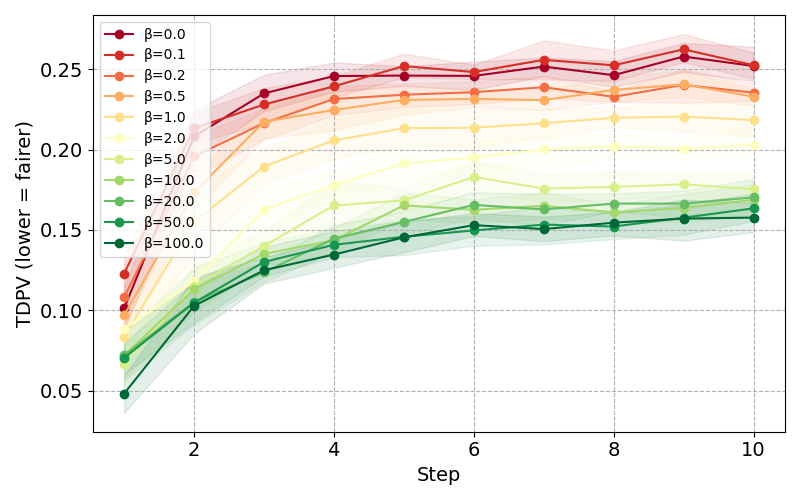}%
    \includegraphics[width=0.49\linewidth]{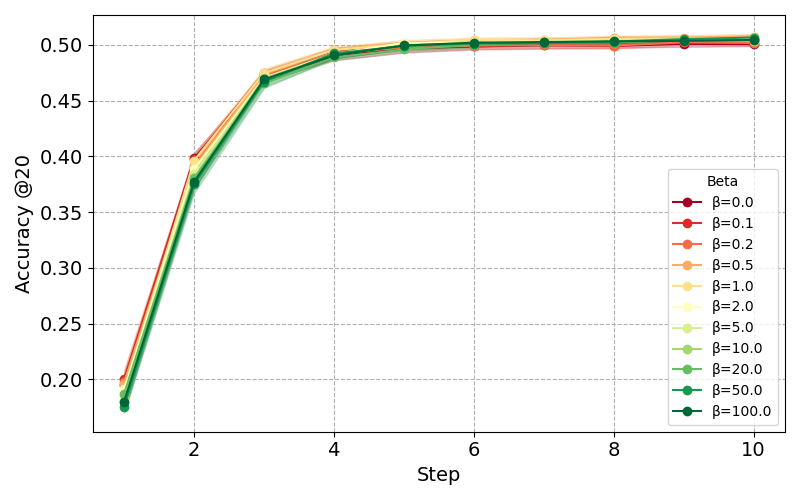}
    \vspace{-0.5em}
    \caption{MetaPath2Vec}
    \label{fig:results_vs_step_metapath}
  \end{subfigure}

  \caption{Evolution of TDPV and Accuracy over training steps for both models. Higher $\beta$ values are in green.}
  \label{fig:combined_results_vs_step}
\end{figure}
% \begin{figure}[t]
%     \centering
%     \includegraphics[width=0.49\linewidth]{Figures/transformer/TDPV_vs_step_ci.png}
%     \includegraphics[width=0.49\linewidth]{Figures/transformer/Acc_vs_Step.png}
%     \caption{Evolution of TDPV and Accuracy over training steps for the transformer model. Higher $\beta$ values are in green.}
%     \label{fig:results_vs_step_transformer}
% \end{figure}
% \begin{figure}
%     \centering
%     \includegraphics[width=0.49\linewidth]{Figures/metapath/TDPV_vs_step_ci.png}
%     \includegraphics[width=0.49\linewidth]{Figures/metapath/Acc_vs_Step.png}
%     \caption{Evolution of TDPV and Accuracy over training steps for the MetaPath2Vec model. Higher $\beta$ values are in green.}
%     \label{fig:results_vs_step_metapath}
% \end{figure}

\begin{table}[t]
  \centering
  \scriptsize
  \setlength{\tabcolsep}{3pt}
  \renewcommand{\arraystretch}{0.9}
  \caption{Accuracy@20 after the final step (Mean ± SE) by group for MetaPath2Vec and Transformer models. Higher $\beta$ value results in a lower TDPV while maintaining accuracy.}
  \begin{tabular}{|l|cc|cc|}
    \hline
    \textbf{Group} & \multicolumn{2}{c|}{\textbf{MetaPath2Vec (\%)}} & \multicolumn{2}{c|}{\textbf{Transformer (\%)}} \\
    \hline
    & \(\beta=0\) & \(\beta=100\) & \(\beta=0\) & \(\beta=100\) \\
    \hline
    \textbf{Overall} & \(50.07 \pm 0.09\) & \(\mathbf{50.45 \pm 0.10}\) & \(\mathbf{28.08 \pm 0.24}\) & \(28.07 \pm 0.21\) \\
    White           & \(\mathbf{52.63 \pm 0.11}\) & \(52.30 \pm 0.13\) & \(\mathbf{29.87 \pm 0.28}\) & \(29.71 \pm 0.20\) \\
    Hispanic        & \(48.80 \pm 0.12\) & \(\mathbf{49.26 \pm 0.12}\) & \(27.91 \pm 0.21\) & \(\mathbf{27.96 \pm 0.26}\) \\
    Asian           & \(45.03 \pm 0.18\) & \(\mathbf{47.43 \pm 0.11}\) & \(23.82 \pm 0.16\) & \(\mathbf{24.16 \pm 0.17}\) \\
    Black           & \(46.40 \pm 0.13\) & \(\mathbf{48.12 \pm 0.10}\) & \(24.81 \pm 0.30\) & \(\mathbf{25.00 \pm 0.20}\) \\
    \hline\hline
    \textbf{TDPV}   & \(25.19 \pm 0.57\) & \(\mathbf{15.77 \pm 0.49}\) & \(21.23 \pm 0.35\) & \(\mathbf{19.62 \pm 0.42}\) \\
    \hline
  \end{tabular}
  \label{tab:accuracy_summary}
\end{table}

\begin{figure}[t]
    \centering
    \includegraphics[width=0.48\linewidth]{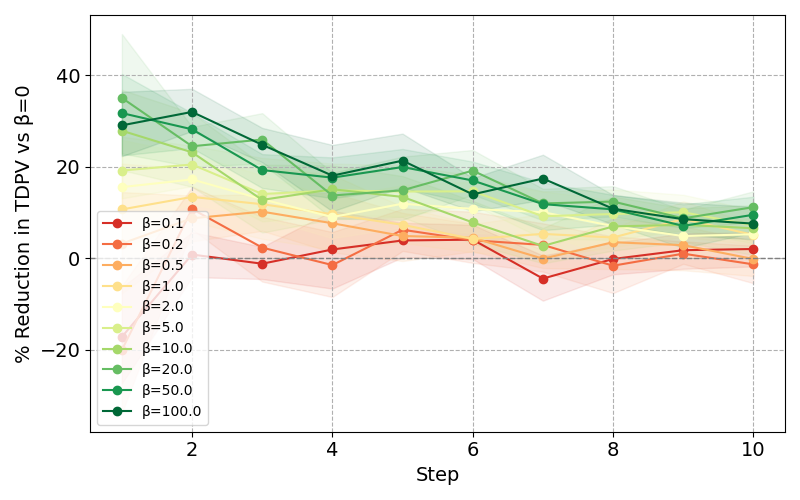}
    % \caption{Percentage reduction in TDPV using FGIS, compared to using $\beta=0$ for the transformer model, with shaded region showing the 95\% bootstrapped confidence interval.}
    % \label{fig:results_pct_reduc_transformer}
% \end{figure}
% \begin{figure}[t]
    \centering
    \includegraphics[width=0.48\linewidth]{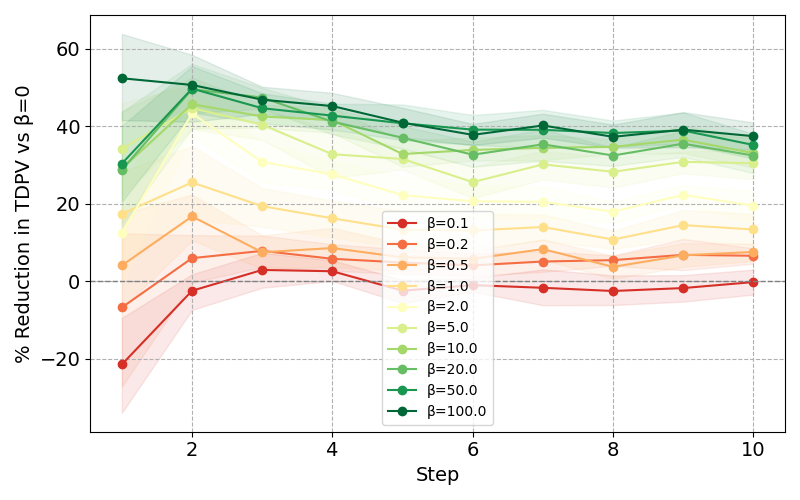}
    \caption{Percentage reduction in TDPV using FGIS, compared to using $\beta=0$ for the transformer model (left) and the MetaPath2Vec model (right), with shaded region showing the 95\% bootstrapped confidence interval.}
    \label{fig:results_pct_reduc}
\end{figure}

\begin{figure*}[t]
  \centering
  \begin{subfigure}{\columnwidth}
    \centering
    \includegraphics[width=0.95\linewidth]{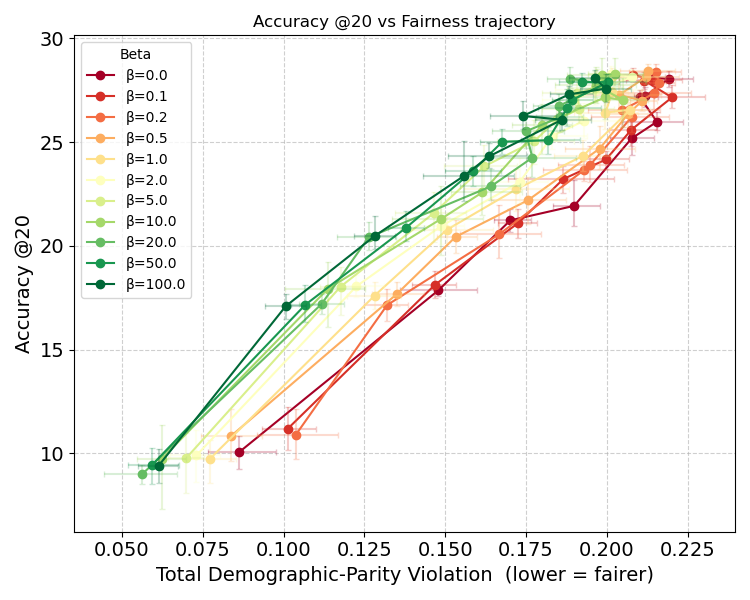}
    \vspace{-0.5em}
    \caption{Transformer}
    \label{fig:results_paret_transformer}
  \end{subfigure}
  \begin{subfigure}{\columnwidth}
    \centering
    \includegraphics[width=0.95\linewidth]{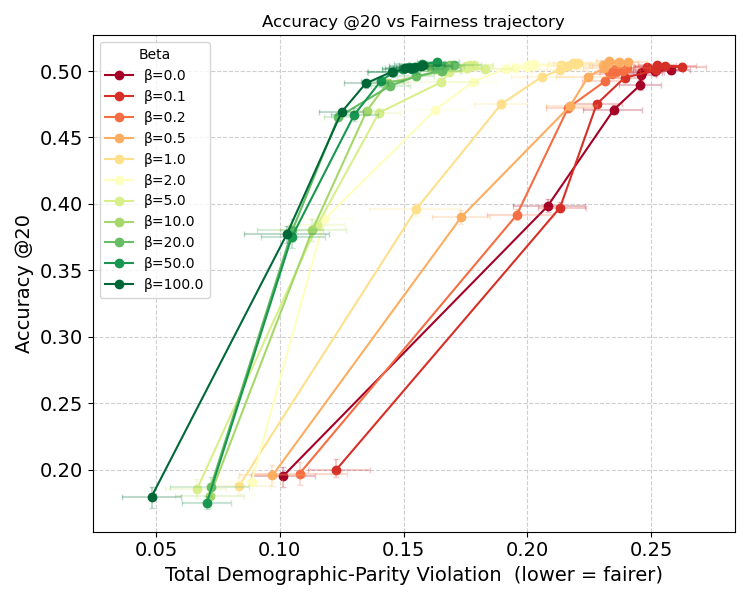}
    \vspace{-0.5em}
    \caption{MetaPath2Vec}
    \label{fig:results_pareto_metapath}
  \end{subfigure}
  
  \caption{Evolution of TDPV and Accuracy@20 over training steps in phase space for both models. The ideal point is top-left. Higher $\beta$ values are in green.}
  \label{fig:combined_results_pareto}
\end{figure*}

% \begin{figure}[t]
%     \centering
%     \includegraphics[width=0.8\linewidth]{Figures/transformer/TDPV_vs_Acc.png}
%     \caption{Evolution of TDPV and Accuracy over training steps in phase space for the transformer model. Ideal point is top-left.  Higher $\beta$ values are in green.}
%     \label{fig:results_paret_transformer}
% \end{figure}
% \begin{figure}[t]
%     \centering
%     \includegraphics[width=0.8\linewidth]{Figures/metapath/TDPV_vs_Acc.png}
%     \caption{Evolution of TDPV and Accuracy over training steps in phase space for the MetaPath2Vec model. Ideal point is top-left.  Higher $\beta$ values are in green.}
%     \label{fig:results_pareto_metapath}
% \end{figure}

\subsection{Alignment with Audit Results}

To establish a baseline and analyze the relative performance of the two selected prediction approaches, we train models with $\beta=0$ (uniform sampling) and $\beta=100$ on the Tarrant County dataset. Table~\ref{tab:accuracy_summary} summarizes the results. We find that MetaPath2Vec is a much stronger model compared to the the transformer-based approach. We attribute this to the difference in tasks. While~\citet{zhang2025colocation} trained to identify new/future visits using colocation networks, the transformer-based approach~\cite{hong2022transformerbaseline} trains on next-location prediction. This indicates the graph-based MetaPath2Vec model benefits significantly from structural information within the training data.

Second, we see that the trends observed in the fairness audit hold even for Tarrant County, with White being the most favored group, followed by Hispanic. This also gives credibility to our SAKM approach used to ground the pseudo-group labels, as it results in clusters with meaningful separation.
Finally, the $\beta=100$ columns show how FGIS improves fairness by bringing these group accuracies closer together without reducing overall accuracy. 
% This is further discussed below. 

% Before analyzing the effect of FGIS, we examine how predictive performance evolves across demographic groups as training progresses with uniform sampling. Figures~\ref{fig:group_evolution_transformer} and ~\ref{fig:group_evolution_metapath} shows the group-wise Acc@20 scores over sampling steps, using SAKM-assigned proxy labels.

% We observe that users labeled as White and Hispanic consistently achieve higher accuracy than those labeled as Black or Asian. This pattern closely mirrors the disparity trends identified in our statewide audit using the MetaPath2Vec model, despite differences in model architecture and geographic scope. These findings reinforce two key points: (1) The SAKM-based proxy labeling produces clusters that meaningfully reflect underlying performance disparities, even without access to true demographic labels; and (2) the observed disparities are not model-specific artifacts, but likely reflect structural imbalances in the data itself.

% This alignment validates the utility of the SAKM proxy labels for group-level fairness evaluation, and provides a consistent baseline against which we evaluate the fairness impacts of our sampling intervention.

\subsection{Impact of FGIS}

To evaluate the effect of FGIS, we run experiments with varying values of the fairness weight parameter $\beta$. A higher value of $\beta$ places more emphasis on equitable sampling when selecting the next batch. 
% We assess the effects of increasing $\beta$ on both predictive accuracy and fairness disparity using the transformer model trained on Tarrant County data.

\subsubsection{Accuracy Impact of Increasing $\beta$:}

Figures~\ref{fig:results_vs_step_transformer}(right) and~\ref{fig:results_vs_step_metapath}(right) presents the evolution of top-20 accuracy over successive sampling steps, where each line corresponds to a different $\beta$ value (with higher $\beta$ shown in green). In early iterations, we observe a small drop in accuracy with higher $\beta$ 
% (e.g., from 8\% at $\beta=0$ to 6.5\% at $\beta=100$ in step 1). 
However, as more users are sampled and the training set grows, the gap in performance rapidly closes. By step 4, all settings converge to a similar accuracy and accuracy starts plateauing, suggesting that fairness-aware sampling does not significantly compromise long-term predictive performance.

\subsubsection{Fairness Impact (TDPV Reduction):}

The effect of FGIS on fairness is illustrated in Figures~\ref{fig:results_vs_step_transformer}(left) and~\ref{fig:results_vs_step_metapath}(left), which shows the change in demographic parity violations (TDPV) over training steps. We find that larger $\beta$ values consistently reduce TDPV, with improvements of up to 30\%  (transformer) and 50\% (MetaPath2Vec) in early iterations and sustained gains of around 10\% (transformer) and 35\% (MetaPath2Vec) by the final step. These trends are further quantified in Figure~\ref{fig:results_pct_reduc}, which shows the percentage reduction in TDPV relative to the baseline ($\beta=0$) along with 95\% bootstrapped confidence intervals. We see that the MetaPath2Vec models benefit significantly more from FGIS, possibly due to their higher predictive power.

\subsubsection{Pareto Analysis of Accuracy-Fairness Tradeoff:}

To visualize the joint behavior of accuracy and fairness, Figure~\ref{fig:combined_results_pareto} plots the trajectory of each $\beta$ configuration in phase space, with Acc@20 on the vertical axis and TDPV on the horizontal axis. Each line traces the model’s performance over time, and colors follow the same scheme as previous figures (red = low $\beta$, green = high $\beta$). The green trajectories clearly Pareto dominate the red ones—achieving lower disparity without sacrificing final accuracy. This suggests that FGIS effectively navigates the fairness-utility tradeoff, yielding substantial disparity reduction with minimal performance cost.

\section{Discussion}

Our experiments show that FGIS delivers substantial early reductions in performance disparity (30-50\% TDPV drop) with almost no long-term accuracy penalty—by the fourth batch all strategies converge to the same Acc@20. This demonstrates that fairness-aware sampling can secure equity gains quickly without sacrificing overall utility.

The fact that SAKM-derived proxy clusters reproduce the statewide MetaPath2Vec audit trends and mirror them under a Transformer in Tarrant County confirms that our census-informed clustering captures the key structural biases. Even without true demographic labels, these proxies enable effective, unsupervised fairness interventions.

As FGIS requires only group counts and periodic accuracy estimates, it can be slotted into any incremental data-collection pipeline with a single tuning parameter $\beta$. Future work might validate SAKM against ground-truth surveys, explore adaptive $\beta$ schedules, and extend to other regions or attributes. By choosing who to collect data from rather than overhauling models, we unlock a lightweight, scalable route to fairer mobility predictions.

\subsubsection{Limitations:}

Our study has certain limitations that we address here. First, while our statewide fairness audit uses a strong graph-based model (MetaPath2Vec), our intervention analysis is restricted to a single county due to computational constraints. While Tarrant County was carefully selected as a representative and demographically diverse subregion, the generalizability of our findings to other geographies or models is an important future task.

Second, our fairness evaluations depend on proxy group labels derived from SAKM. While these proxies align with census distributions and reflect known disparities, they are not a substitute for ground-truth demographic attributes.

Finally, our sampling strategy introduces a fairness-utility tradeoff via a manually selected hyperparameter $\beta$. Future work could explore adaptive approaches that dynamically balance this tradeoff during data acquisition.

\section{Conclusion}

We present the first fairness audit of large-scale individual-level next-location prediction and propose a lightweight intervention for reducing group disparities. Our audit reveals consistent performance gaps across geographic, racial and ethnic lines, even in high-performing models like MetaPath2Vec.
To address this, we introduce Fairness-Guided Incremental Sampling (FGIS), a data-first intervention which steers data collection toward underrepresented or underperforming groups. Using proxy labels from Size-Aware K-Means (SAKM), our method achieves up to 40\% disparity reduction with minimal accuracy loss.

Together, these results underscore the risks of overlooking fairness in mobility prediction and demonstrate that simple, model-agnostic sampling strategies can yield meaningful equity improvements without requiring access to sensitive user data or changes to model architecture. In future work, we plan to extend our evaluation to additional geographies to assess generalizability across regions.

\bibliography{main}

\newpage
% \input{ReproducibilityChecklist/ReproducibilityChecklist.tex}
\appendix

\section{Extended Fairness Audit Results}

We present some additional results from our fairness evaluation of the MetaPath2Vec model on the full Texas dataset here.

\subsection{Geographic Disparities}
\begin{figure}
    \centering
    \includegraphics[width=0.49\linewidth]{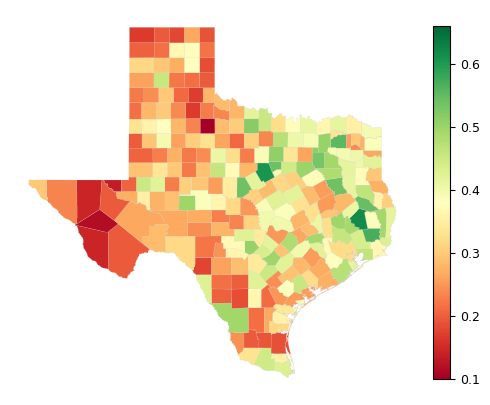}
    \includegraphics[width=0.49\linewidth]{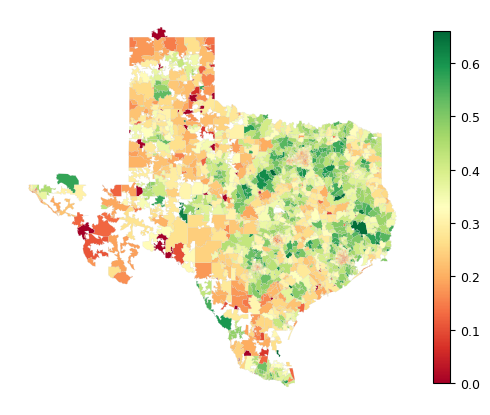}
    \caption{Top-20 prediction accuracy over a one-week lookahead period across Texas counties and ZCTAs.}
    \label{fig:maps}
\end{figure}

To assess spatial variation, we compute the average prediction accuracy at both the county and ZIP Code Tabulation Area (ZCTA) levels, based on each user's inferred home region. The geographic distribution of accuracy is shown in Figure~\ref{fig:maps}.

We observe that prediction accuracy tends to be higher in northeastern Texas, particularly in areas with greater population density and more abundant data coverage. In contrast, western Texas, which is more sparsely populated, generally exhibits lower predictive performance.

\begin{figure}
    \centering
    \includegraphics[width=0.49\linewidth]{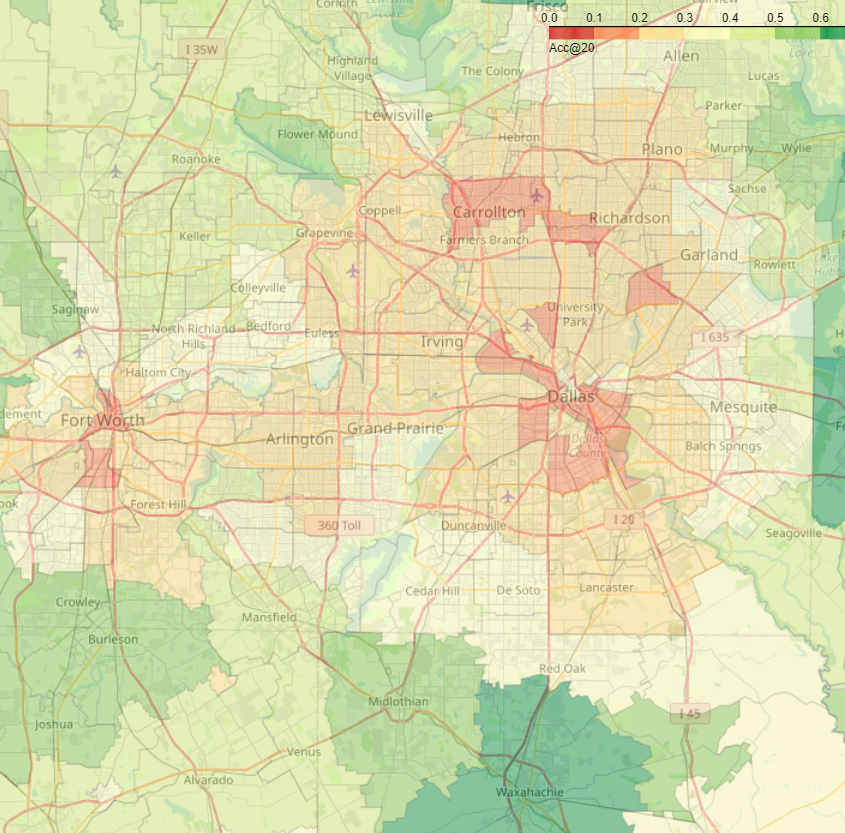}
    \includegraphics[width=0.49\linewidth]{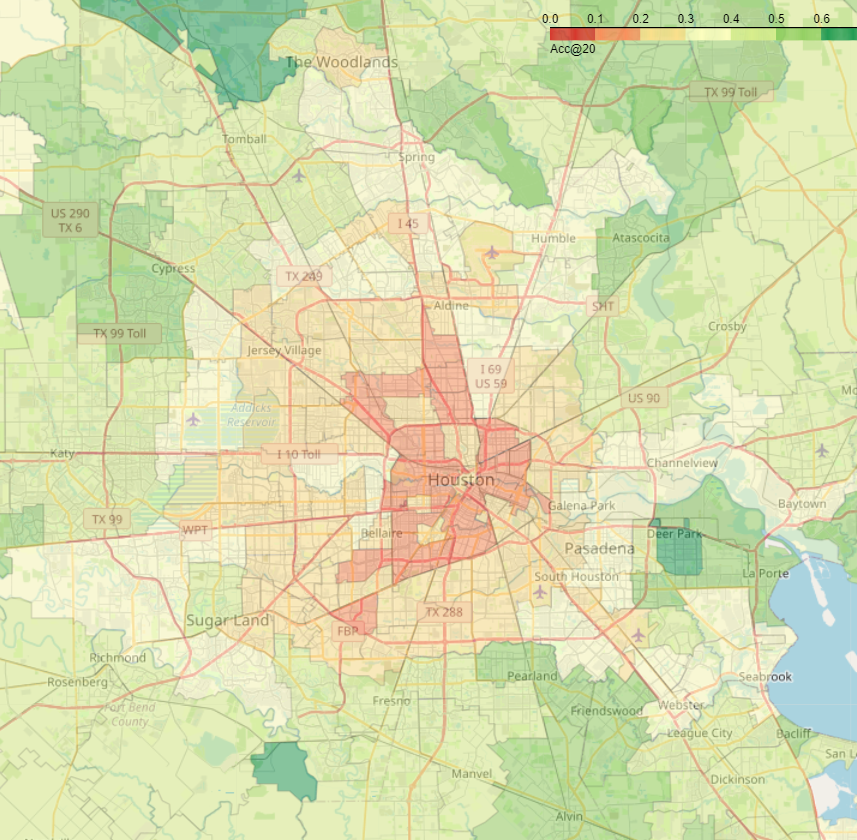}
    \caption{Differences in urban vs. suburban prediction accuracy in Dallas (left) and Houston (right).}
    \label{fig:cities}
\end{figure}

However, the relationship between population density and model accuracy is not strictly monotonic. In major metropolitan areas such as Dallas and Houston (Figure~\ref{fig:cities}), suburban regions often achieve higher accuracy than central urban zones. We attribute this to the increased difficulty of predicting mobility in dense urban areas with a high concentration of nearby POIs, which introduces greater ambiguity despite larger data volumes.

\subsection{Racial/Ethnic Disparities}
\begin{figure}[t]
    \centering
    \includegraphics[width=0.8\linewidth]{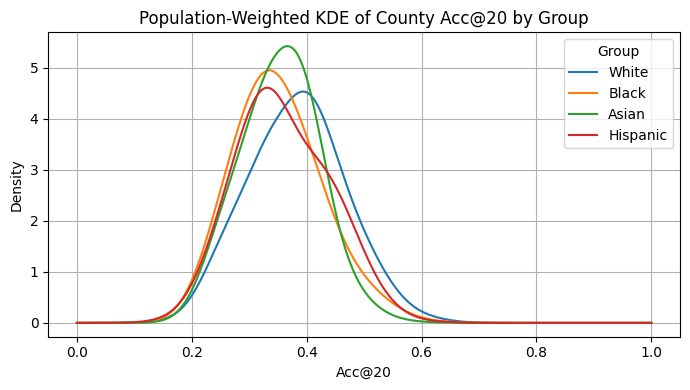}
    \includegraphics[width=0.8\linewidth]{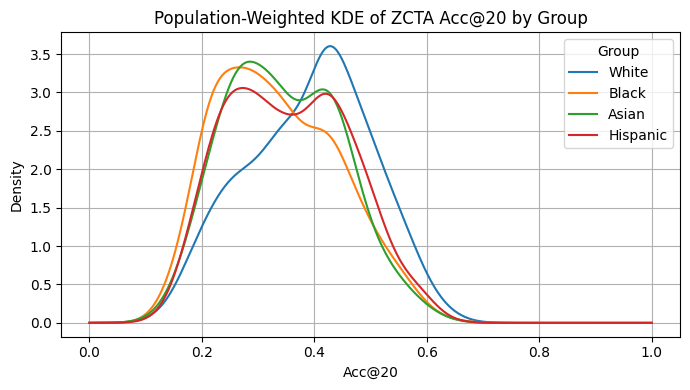}
    \caption{Kernel density estimates of group-level prediction accuracy distributions. Note: density reflects intra-group distribution, not relative group sizes.}
    \label{fig:kde}
\end{figure}

In addition to the results in the main text, Figure~\ref{fig:kde} presents kernel density estimates of the accuracy distributions across the population of each group, showing the density of users at each accuracy level based on racial/ethnic groups.

\begin{algorithm}[t]
\caption{SAKM: Size-Aware K-Means with Permutation Search (Main)}
\label{alg:sakm_main}
\textbf{Input}: Data $X \in \mathbb{R}^{N \times d}$, target proportions $\boldsymbol{\pi}$, clusters $k$\\
\textbf{Parameter}: Step-size $\eta$, tolerance $\tau$, max iterations $T$, restarts $n_{\text{init}}$\\
\textbf{Output}: Best cluster assignments $\mathbf{z}$ and centroids $\{\mu_j\}$
\begin{algorithmic}[1]
\STATE $\mathcal{L}_{\text{best}} \gets \infty$
\FOR{$s = 1$ to $n_{\text{init}}$}
    \STATE Initialize centroids $\{\mu_1, \dots, \mu_k\}$
    \FOR{each permutation $\boldsymbol{\pi}'$ of $\boldsymbol{\pi}$}
        \STATE $\mathbf{z}, \{\mu_j\}, \mathcal{L} \gets \textsc{RunSAKMInnerLoop}($
        \STATE \hspace{1.5em} $X, \{\mu_j\}, \boldsymbol{\pi}', \eta, \tau, T)$
        \IF{$\mathcal{L} < \mathcal{L}_{\text{best}}$}
            \STATE Save $\mathbf{z}, \{\mu_j\}, \mathcal{L}_{\text{best}} \gets \mathcal{L}$
        \ENDIF
    \ENDFOR
\ENDFOR
\STATE \textbf{return} best $\mathbf{z}, \{\mu_j\}$
\end{algorithmic}
\end{algorithm}

\begin{algorithm}[t]
\caption{\textsc{RunSAKMInnerLoop}: Assignment and Multiplier Updates}
\label{alg:sakm_inner}
\textbf{Input}: Data $X$, initial centroids $\{\mu_j\}$, target proportions $\boldsymbol{\pi}'$, step-size $\eta$, tolerance $\tau$, iterations $T$\\
\textbf{Output}: Final assignments $\mathbf{z}$, centroids $\{\mu_j\}$, objective $\mathcal{L}$
\begin{algorithmic}[1]
\STATE Initialize $\lambda_j \gets 0$ for all $j$
\FOR{$t = 1$ to $T$}
    \FOR{each user $x_i$}
        \STATE Assign $z_i \gets \arg\min_{j} \|x_i - \mu_j\|^2 + \lambda_j$
    \ENDFOR
    \FOR{each cluster $j$}
        \STATE Update $\mu_j \gets$ mean of $\{x_i : z_i = j\}$
    \ENDFOR
    \FOR{each cluster $j$}
        \STATE Let $n_j \gets \#\{i : z_i = j\}$
        \STATE $\lambda_j \gets \lambda_j + \eta \cdot \left( \frac{n_j}{N} - \pi_j' \right)$
    \ENDFOR
    \IF{centroids shift $< \tau$}
        \STATE \textbf{break}
    \ENDIF
\ENDFOR
\STATE Compute $\mathcal{L} \gets \sum_i \|x_i - \mu_{z_i}\|^2$
\STATE \textbf{return} $\mathbf{z}, \{\mu_j\}, \mathcal{L}$
\end{algorithmic}
\end{algorithm}
\section{Size-Aware K-Means for Proxy Demographic Grounding}

Our fairness-aware sampling strategy assumes access to group membership labels, but individual-level demographics are not observed in our dataset. To address this, we construct \textit{proxy demographic labels} based on available census priors at the ZCTA level. These priors give us coarse estimates of racial group proportions, which we use to guide an unsupervised clustering process.

Rather than assigning users to groups arbitrarily or uniformly, we seek a principled partitioning that reflects the demographic composition of the local population. To this end, we embed users into a latent space derived from their mobility trajectories and cluster them into $k$ groups, one per racial category. While these clusters may not align exactly with true group identities, they offer a structure that is both data-driven and demographically grounded. We provide some context from related work for this approach here.

\subsection{Size-based Clustering}
Existing attempts to control cluster sizes either force \emph{equal} cardinality or impose only broad capacity limits, leaving no practical tool for matching an arbitrary quota vector~$\boldsymbol{\pi}$.  Constrained and “balanced” $k$-means variants guarantee $|C_{1}|=\!\dots=\!|C_{k}|$ via repeated Hungarian assignment~\citep{Bradley2000,malinen2014balanced}, while flow-based approaches minimize a cost plus a penalty for deviation from \emph{uniform} sizes~\citep{lin2019balanced}.  Capacity-constrained $k$-means from operations research merely caps the load per cluster~\citep{geetha2009improved}, and exact-quota mixed-integer or conic formulations scale only to a few thousand points~\citep{rujeerapaiboon2019size}.  In contrast, our \textbf{Size-Aware $k$-Means} keeps Lloyd-style updates but adds dual Lagrange steps, efficiently steering each cluster toward \emph{any} prescribed proportion vector.

\subsection{Algorithm}

We implement a modified clustering algorithm, \textit{Size-Aware K-Means (SAKM)}, which extends standard $k$-means to enforce user-defined cluster size constraints. Given a target group distribution $\boldsymbol{\pi} = (\pi_1, \dots, \pi_k)$, SAKM aims to produce clusters of sizes close to $\pi_g \cdot N$, where $N$ is the total number of users.

SAKM introduces a Lagrangian penalty in the assignment step, so the cost of assigning a point $x_i$ to cluster $g$ becomes:
\begin{align}
    \text{cost}(x_i, g) = \|x_i - \mu_g\|^2 + \lambda_g \label{eq:sakm-cost}
\end{align}
where $\mu_g$ is the centroid of cluster $g$ and $\lambda_g$ is a Lagrange multiplier that penalizes deviations from the target size. These multipliers are updated iteratively as:
\begin{align}
\lambda_g \leftarrow \lambda_g + \eta \cdot \left( \frac{n_g}{N} - \pi_g \right) \label{eq:sakm-largrange}
\end{align}
where $n_g$ is the current number of points in cluster $g$. 
To resolve the label ambiguity and improve convergence quality, we run the optimization over all permutations of the target proportions $\boldsymbol{\pi}$ and select the clustering with the lowest objective (inertia). This permutation search is motivated by a key challenge in centroid initialization: if a randomly initialized centroid is far from the true region of the corresponding size, the resulting cluster may fail to attract the intended mass of points. By evaluating all $k!$ permutations of the size targets, SAKM increases the likelihood that size constraints are matched with semantically meaningful partitions.

While the worst-case complexity grows factorially with the number of groups, this remains tractable in our setting where $k=4$. For larger $k$, the cost could be reduced using combinatorial assignment methods such as the Hungarian algorithm with $\mathcal{O}(k^3)$ complexity, which we explored but do not report in this paper. 
% For our use case, exhaustive permutation search does not incur a significant runtime overhead.

The full algorithm is presented in Algorithm~\ref{alg:sakm_main}.

\subsection{SAKM Calibration Results}
\begin{figure}
    \centering
    \includegraphics[width=0.49\linewidth]{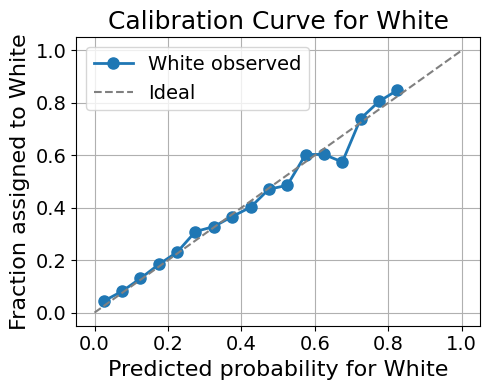}
    \includegraphics[width=0.49\linewidth]{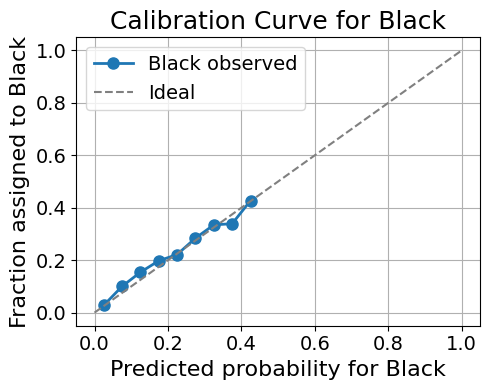}
    \includegraphics[width=0.49\linewidth]{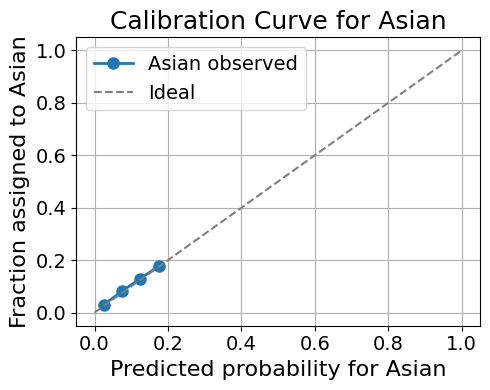}
    \includegraphics[width=0.49\linewidth]{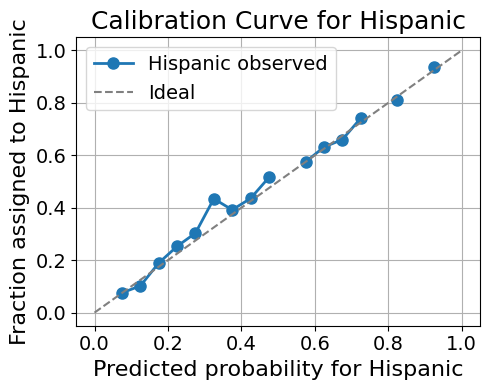}
    \caption{Calibration curves for the different demographic groups.}
    \label{fig:SAKM-calibration}
\end{figure}
We used the exhaustive Size-Aware K-Means algorithm to generate proxy labels for users in Tarrant County (used in our main experiments), using the ZCTA level census data to set cluster sizes. We performed 50 maximum K-means iterations and 2 random initializations.
To validate the SAKM output, we compare the resulting cluster proportions to the original census-derived target distribution. Figure~\ref{fig:SAKM-calibration} shows the calibration curves for all groups, after filtering for ZCTAs with fewer than 10 users to reduce noise. SAKM consistently yields clusters that match the target distribution within a small margin, confirming that our proxy labeling aligns with the intended demographic structure. This grounding enables meaningful downstream evaluation and intervention in our group-aware prediction tasks.

\begin{figure*}[t]
  \centering
  \begin{subfigure}[b]{0.48\textwidth}
    \includegraphics[width=\linewidth]{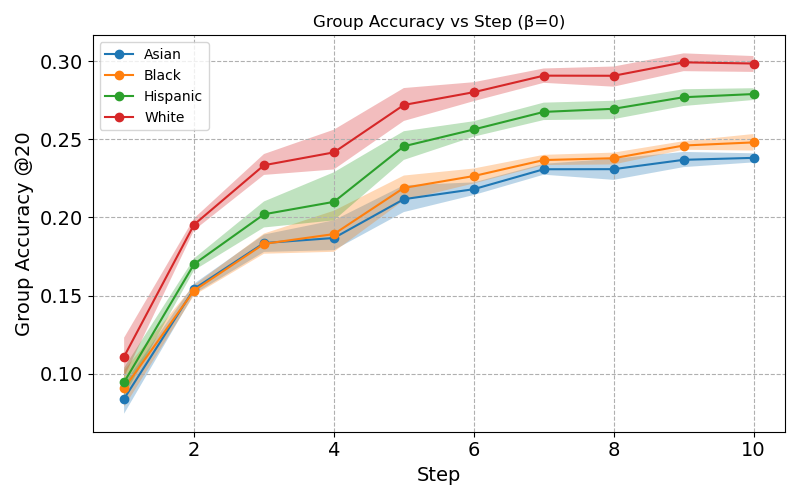}
    \caption{$\beta=0$: White and Hispanic groups converge to higher accuracy.}
    \label{fig:transformer_beta0}
  \end{subfigure}
  \hfill
  \begin{subfigure}[b]{0.48\textwidth}
    \includegraphics[width=\linewidth]{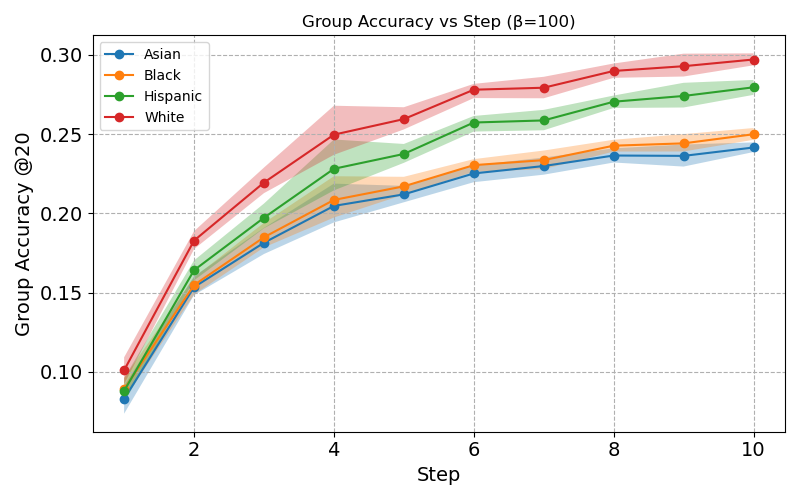}
    \caption{$\beta=100$: Fairness‐aware sampling reduces disparity.}
    \label{fig:transformer_beta100}
  \end{subfigure}
  \caption{Evolution of group accuracy over training steps for the \textbf{transformer} model.}
  \label{fig:group_evolution_transformer}
\end{figure*}

\section{Experimental Details}
\subsection{Metpath2Vec}
% \ak{@Hanyu Here we will go into detail about the metapath model. HZ: Updated. Let me know how detailed you’d like this part to be, or if the language looks good to you.}

We adapt the MetaPath2Vec architecture introduced by \cite{dong2017metapath} and follow the evaluation framework of \cite{zhang2025colocation}, which leverages user–POI visitations and user-user colocations to construct heterogeneous networks for predicting consumer visits. In our implementation, we construct a user-POI visitation network, where nodes represent users or POIs, and edges indicate observed visits. We then generate Meta-path-guided random walks (e.g., user–POI)  sequences to capture structural and semantic proximity, enabling the model to learn user and POI embeddings ($d_{users}$ and $d_{pois}$) via a skip-gram model. We then compute similarities between POIs using Euclidean or cosine distance in the embedding space.

% The model is trained on visit data prior to March 15, 2021 and evaluated on the holdout period from March 15 to April 15, 2021. 
For each user in the test set, we compute the $k$ closest POIs based on embedding distances. These $k$ POIs are treated as predicted next visits. A hit is recorded if the user visits at least one of the predicted POIs during the first week of the holdout period. We evaluate hit rate using $k$ = 20.

\begin{table}[ht]
\centering
\caption{MetaPath2Vec Model Settings and Hyperparameters}
\begin{tabular}{l c}
\hline
\textbf{Setting} & \textbf{Value} \\
\hline
Walks per Node & 10 \\
Walk Length & 100 \\
Embedding Dimension ($d$) & 128 \\
Neighborhood Size  & 7 \\
Negative Samples & 5 \\
\hline
\end{tabular}
\label{tab:metapath_params}
\end{table}

\subsection{Transformer Encoder}
% While our fairness audit used the MetaPath2Vec model for statewide evaluation, we were unable to access its training code to support our proposed intervention. As an alternative, we implement a transformer-based next-location prediction model to evaluate the effects of our fair sampling method.

Transformer architectures are widely used in sequence modeling and represent a strong alternative for mobility prediction. 
In particular, we adapt a transformer-based architecture introduced by ~\citet{hong2022transformerbaseline}, originally designed to improve next-location prediction by jointly modeling travel sequences and travel modes. Their model employs a transformer encoder to capture spatiotemporal dependencies in mobility histories, with an auxiliary head trained to predict the next travel mode alongside the next location. In our implementation, we retain the core sequence modeling architecture but omit the auxiliary travel mode prediction to reduce complexity and accommodate datasets without modality labels.
We add periodic embeddings for the time to next location during training, and compute user embeddings (of size $d_{user}$) using the approximate home location coordinates. For the transformer encoder, the inputs are sequences of POIs and their timestamps. The embedding layer (of size $d_{base}$) is computed by embedding the POI IDs and adding a time-of-day embedding, before applying sinusoidal position encoding. 
The next POI prediction is computed by concatenating the user embedding, time-to-next embedding and sequence embedding (from the transformer encoder), and computing logits over all possible POIs after a feedforward layer. The resulting probability distribution gives us the top-k POI prediction, which is then used to compute downstream metrics.

% While transformer architectures offer strong representational capacity, their scalability remains a practical limitation in large-scale mobility settings. The full Texas dataset comprises over 5 million trajectories and more than 500{,}000 unique POIs, rendering end-to-end transformer training computationally infeasible for independent research groups. Consequently, we restrict our intervention analysis to Tarrant County, a geographically and demographically diverse subregion where the data volume is sufficiently constrained to support tractable experimentation.

Table~\ref{tab:transformer_hparams} shows the configuration for the transformer model used. 
\begin{table}[h!]
    \centering
    \begin{tabular}{lr|lr}
        \hline
        \textbf{Setting} & \textbf{Value} & \textbf{Parameter} & \textbf{Value} \\
        \hline
        LR & 0.001 & \# Layers & 4 \\
        LR decay& 1e-6 & \# Heads & 4 \\
        LR Warmup & 2 & Feedforward & 512 \\
        ES Patience & 2 & Base Emb. ($d_{\text{base}}$) & 256 \\
        ES LR drop & 0.33 & User Emb. ($d_{\text{user}}$) & 8 \\
         &  & FC Dropout & 0.1 \\
        \hline
    \end{tabular}
    \caption{Transformer Model Settings and Hyperparameters}
    \label{tab:transformer_hparams}
\end{table}

\subsection{Tarrant County Subset}

Our experiments require training and evaluating models many times—both to probe the fairness–accuracy tradeoff under different sampling conditions and to compute confidence intervals via bootstrapping. To keep these repeated runs tractable, especially for the Transformer model, we restrict our analysis to a subset of the Texas dataset focused on a single county.

We select this county for our case study using a simple two-step filter. First, we compute a population-weighted disparity score for each county, defined as the product of inter-group accuracy variance and county population under the original state-level MetaPath2Vec model. This score highlights regions where predictive disparities are both large and demographically consequential. Second, we limit to counties with fewer than 200,000 total trajectories to ensure practical training costs.

Tarrant County emerged as the top candidate, exhibiting significant fairness gaps under the statewide MetaPath2Vec model while remaining computationally manageable for repeated analysis. After removing users and POIs outside of Tarrant County, the final filtered dataset contained 170,000 user trajectories and 39,000 unique POIs.

\subsection{Implementation Details}
All experiments were conducted on a shared university compute cluster. Each run was allocated a single GPU with 8GB of VRAM and 32GB of system RAM. 
Each training iteration sampled 1,000 users, and the models were trained over 10 such iterations, yielding a total of 10,000 unique sampled users per experiment. After sampling and adding each new batch, a new model was initialized and trained from scratch. A full experiment consisting of 10 iterations typically completed within 24 hours. To assess statistical significance and reduce variance, we repeat each configuration with 10 random seeds and report aggregate metrics. The transformer model for predicting next location for Tarrant county had 22.9M trainable parameters.%22983056

The underlying dataset spans the period from January 1 to April 15, 2021. The training split spans the period from January 1 to March 15, and models were evaluated on a held-out test period from March 15 to March 22 (one week). For the transformer model, we further divided the training set into two parts for training and validation: the model was trained on data from January 1 to March 1 and validated on sequences from March 1 to March 15. Model selection and early stopping were based on validation accuracy, with learning rate decay triggered by validation plateaus.

This evaluation protocol ensured a consistent lookahead period of one week for computing the Acc@20 metric, aligning with the evaluation framework used in our earlier fairness audit.

\section{Additional Results}

% \begin{figure}
%     \centering
%     \includegraphics[width=0.98\linewidth]{Figures/transformer/Acc_beta_0.png}
%     \caption{Evolution of Group Accuracy over training steps for $\beta=0$ for the transformer model. White and Hispanic groups converge to higher accuracy.}
%     \label{fig:group_evolution_transformer}
% \end{figure}
% \begin{figure}
%     \centering
%     \includegraphics[width=0.98\linewidth]{Figures/transformer/Acc_beta_100.png}
%     \caption{Evolution of Group Accuracy over training steps for $\beta=100$ for the transformer model. The addition of fairness-aware sampling reduces disparity by bringing the group accuracies closer together.}
%     \label{fig:group_evolution_transformer_100}
% \end{figure}
% \begin{figure}
%     \centering
%     \includegraphics[width=0.98\linewidth]{Figures/metapath/Acc_beta_0.png}
%     \caption{Evolution of Group Accuracy over training steps for $\beta=0$ for the MetaPath2Vec model. White and Hispanic groups converge to higher accuracy.}
%     \label{fig:group_evolution_metapath}
% \end{figure}
% \begin{figure}
%     \centering
%     \includegraphics[width=0.98\linewidth]{Figures/metapath/Acc_beta_100.png}
%     \caption{Evolution of Group Accuracy over training steps for $\beta=0$ for the MetaPath2Vec model. The addition of fairness-aware sampling reduces disparity by bringing the group accuracies closer together.}
%     \label{fig:group_evolution_metapath_100}
% \end{figure}

\begin{figure*}[t]
  \centering
  \begin{subfigure}[b]{0.48\textwidth}
    \includegraphics[width=\linewidth]{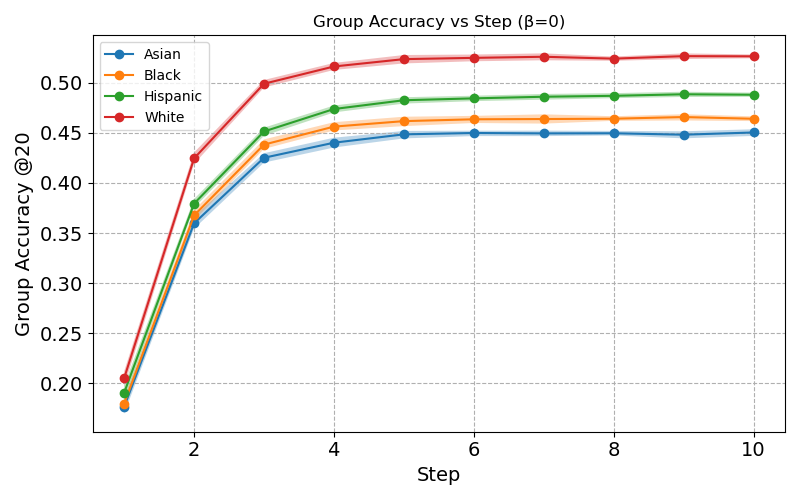}
    \caption{$\beta=0$: White and Hispanic groups converge to higher accuracy.}
    \label{fig:metapath_beta0}
  \end{subfigure}
  \hfill
  \begin{subfigure}[b]{0.48\textwidth}
    \includegraphics[width=\linewidth]{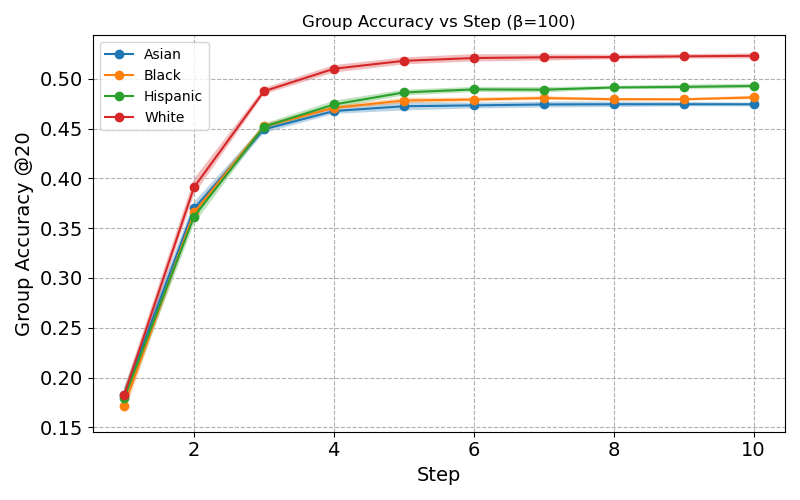}
    \caption{$\beta=100$: Fairness‐aware sampling reduces disparity.}
    \label{fig:metapath_beta100}
  \end{subfigure}
  \caption{Evolution of group accuracy over training steps for the \textbf{MetaPath2Vec} model.}
  \label{fig:group_evolution_metapath}
\end{figure*}

We examine how predictive performance evolves across demographic groups as training progresses with uniform sampling. Figures~\ref{fig:transformer_beta0} and ~\ref{fig:metapath_beta0} show the group-wise Acc@20 scores over sampling steps, using SAKM-assigned proxy labels.

We observe that users labeled as White and Hispanic consistently achieve higher accuracy than those labeled as Black or Asian. This pattern closely mirrors the disparity trends identified in our statewide audit using the MetaPath2Vec model, despite differences in model architecture and geographic scope. These findings reinforce two key points: (1) The SAKM-based proxy labeling produces clusters that meaningfully reflect underlying performance disparities, even without access to true demographic labels; and (2) the observed disparities are not model-specific artifacts, but likely reflect structural imbalances in the data itself.

This alignment validates the utility of the SAKM proxy labels for group-level fairness evaluation, and provides a consistent baseline against which we evaluate the fairness impacts of our sampling intervention.

Then, Figures~\ref{fig:transformer_beta100} and~\ref{fig:metapath_beta100} show this same evolution with a high $\beta$ value ($\beta=100$). 

\end{document}